# Convolution with Logarithmic Filter Groups for Efficient Shallow CNN

Tae Kwan Lee    Wissam J. Baddar    Seong Tae Kim    Yong Man Ro

School of Electrical Engineering, KAIST, Republic of Korea

{ltk010203, wisam.baddar, stkim4978, ymro}@kaist.ac.kr

## Abstract

*In convolutional neural networks (CNNs), the filter grouping in convolution layers is known to be useful to reduce the network parameter size. In this paper, we propose a new logarithmic filter grouping which can capture the nonlinearity of filter distribution in CNNs. The proposed logarithmic filter grouping is installed in shallow CNNs applicable in a mobile application. Experiments were performed with the shallow CNNs for classification tasks. Our classification results on Multi-PIE dataset for facial expression recognition and CIFAR-10 dataset for object classification reveal that the compact CNN with the proposed logarithmic filter grouping scheme outperforms the same network with the uniform filter grouping in terms of accuracy and parameter efficiency. Our results indicate that the efficiency of shallow CNNs can be improved by the proposed logarithmic filter grouping.*

## 1. Introduction

Recently, convolutional neural networks (CNNs) have shown state-of-the-art performance in various classification tasks [1-6], including face recognition [6], facial expression recognition [1, 3], and object classification (e.g. ILSVRC [4, 5, 7]). The increase in performance is largely due to the increased non-linearity in the model and abstractions that allow representation of more complex objects or classes [7].

In constrained conditions, such as embedded systems or mobile devices, networks with smaller parameters are needed due to the limitations of memory and computing power [8]. Therefore, having a CNN classification model small in size, and robust in performance can save memory, and energy in mobile applications. Recently, it has been shown that a reasonably good performance can be achieved with shallower networks for smaller classification tasks (with small number of outcomes) [9, 10]. Nonetheless, model parameters and computational complexity could still be improved in shallower networks.

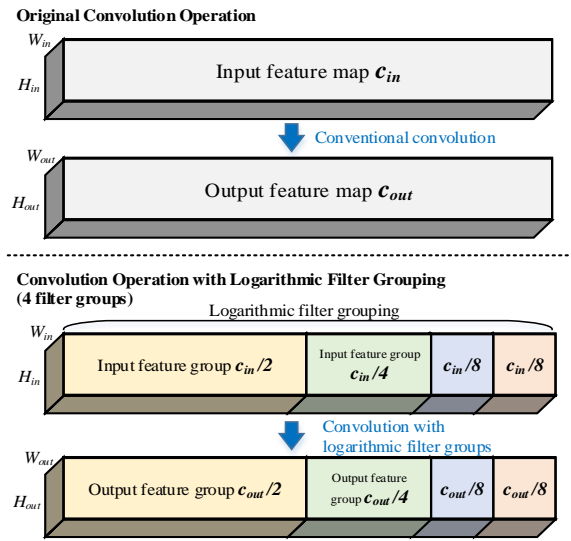

Figure 1: Overview of the proposed nonlinear logarithmic filter grouping in a convolution layer. $H_{in}$, $W_{in}$, $H_{out}$, $W_{out}$ denote the input and output feature map size. $c_{in}$ is total number of input channels, $c_{out}$ is that of output channels for the convolution layer.

Several research efforts have recently shown that the parameters in the CNNs could be reduced while maintaining the performance [8, 11-15]. Some approaches make use of the redundancies in spatial dimensions of filters by factorization of convolutions or low-rank approximations, etc. [11, 13-15]. Another approach introduces the concept of filter grouping [12]. The filter grouping divides a full convolution layer into smaller convolutions with uniform channel size which in sum have the same input and output feature dimensions compared to the full convolution [12]. The method reduces model parameters in deep CNNs while maintaining the performance and enhancing the computation speed [12]. [8] can be seen as an extreme version of [12], where the standard convolution is decomposed into depth-wise and point-wise convolutions. Such methods remove network redundancies in CNNs effectively. However, applying filter grouping in [12] directly to shallow CNNs could degrade the performance of the classification. Moreover, we cannot guarantee whether the uniform filter grouping successfully reflects the nonlinear nature within shallow CNNs.

In this paper, we propose a novel logarithmic filter grouping for a shallow CNN model in general classification tasks (see Fig. 1). Our network develops the proposed logarithmic filter grouping and residual identity connections [2, 7] to reduce parameters in shallow networks while minimizing the performance loss (classification performance loss). The proposed logarithmic filter grouping is devised considering the nonlinear characteristic of filters which resembles the human perception of physical stimuli [16]. Identity connections are incorporated with the proposed filter grouping to encode residual mappings in the shallow CNNs. We show that our compact CNN model with the proposed logarithmic filter grouping shows better classification performance and improved parameter efficiency compared to the uniform filter grouping. The contributions of this paper can be summarized by the following:

1. We propose a new filter grouping which introduces the concept of nonlinear logarithmic filter grouping. The logarithmic filter grouping is devised based on the filters' nonlinear characteristic. By using the proposed filter grouping, the parameters in convolution layers can be reduced while improving classification performance compared to the uniform filter grouping. Further residual identity shortcut is employed to incorporate with the proposed filter grouping for building efficient shallow CNN model.

2. We devise a compact CNN for general classification tasks based on the proposed filter grouping. The model has smaller parameters compared to a baseline compact CNN model with the same depth. At the same time, the proposed CNN architecture with the proposed filter grouping minimizes the performance loss compared to the baseline compact model on different classification tasks (Facial expression recognition using Multi-Pie dataset [17] and object classification using CIFAR-10 dataset [18]).

## 2. Related Work

### 2.1. Hierarchical Filter Groups

Typical convolution filters in CNNs have full connections between the input and output feature maps. If the input feature map has $c_{in}$ channels and the output feature map has $c_{out}$ channels, the filter dimension is $h \times w \times c_{in} \times c_{out}$. This means that the height of the filter is $h$, the width is $w$, the channel depth is $c_{in}$, and there are $c_{out}$ filters of corresponding shapes.

The work in [12] applies filter groups manually to disconnect the connectivity between the input and output feature maps. For example, if $n$ filter groups are applied, $n$ uniform filter groups with $c_{out}/n$ filters are used. Each filter group has a dimension of $h \times w \times c_{in}/n$, i.e. total filter dimension becomes $h \times w \times c_{in}/n \times c_{out}$. Total parameters required for this convolution layer is $n$ times smaller than that of the original full convolution layer.

The degree of grouping $n$ is also reduced by half as the network goes deeper. This 'root topology' exploits the idea that deeper layers need more filter dependencies, such that simpler features are combined to produce more complex features. [12] uses this hierarchical filter group concept to reduce parameters in deep convolutional networks, yet maintaining the performance.

### 2.2. Residual Network

Residual network is an example of a very deep neural network. When a neural network becomes deeper, vanishing gradients problem arises which drops the performance of the network [7]. Layers in residual networks are composed of residual module $F(\cdot)$ and a skip connection bypassing $F(\cdot)$. The identity skip connection allows efficient back-propagation of errors, and resolves the vanishing gradient problem [2]. Another interpretation explains that residual networks behave like ensembles of relatively shallow networks [19]. This ensemble effect is possible due to this identity skip connections [19]. The authors in [20] used a similar concept to train residual networks efficiently. Apart from conveying gradients effectively, identity skip connections have another important role of encoding residual mapping. A residual mapping can be defined by

$$F(x) = H(x) - x \qquad (1)$$

where $x$ denotes the input, $F(x)$ denotes some arbitrary sequence of convolutions, batch normalization [21], and activation functions such as ReLU [22] on input. $H(x)$ is the desired underlying residual mapping. Rearranging the equation we get

$$H(x) = F(x) + x \qquad (2)$$

As equation (2) shows, the identity skip connection allows the encoding of residual mapping. The authors in [2, 7] showed through careful experiments that residual mapping with the identity skip connection is better than plain network in both training easiness and performance. Identity skip connections are utilized in our model to take advantage of the effectiveness of residual mapping. However, because the proposed network is shallow, we assume that the vanishing and exploding gradient problems (addressed in deep CNNs) are insignificant. Our experiments show that residual mapping is not only effective in deep networks, but also useful for enhancing

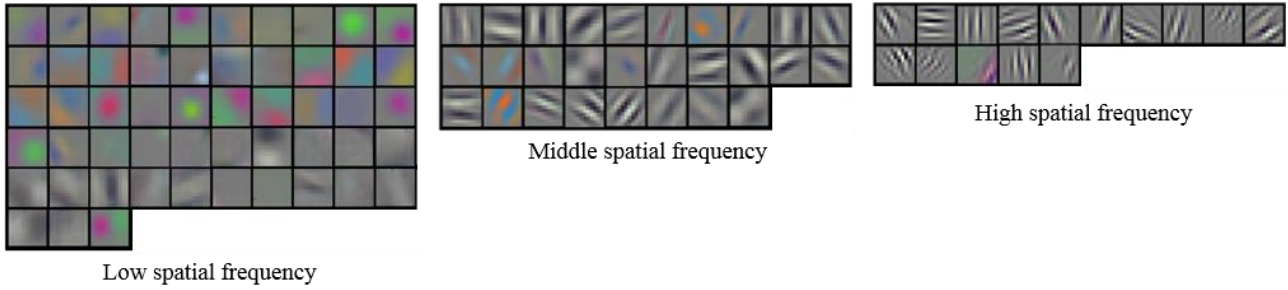

Figure 2: Filter distribution of the first convolutional layer of AlexNet in the viewpoint of spatial frequency [23]. Note that CNN filters are not distributed uniformly and they could not be grouped linearly into low, middle and high spatial frequency filter categories. Among 96 filters in the first convolution layer, the number of filters of each spatial frequency filter category is approximately 53, 28, and 15, which shows a logarithmic like distribution with an approximate base number 2.

the performance of shallower networks.

### 2.3. Bottleneck Architecture

In [5, 7] bottleneck architecture is used to reduce the computational complexity in convolution layers. 1×1 convolution is used to reduce the channel depth before the convolution, and to expand the channel after the convolution. This way fewer parameters are used for convolution layers. A similar idea is used in the proposed network, but in a reversed manner. In the proposed network, 1×1 convolution is used to increase the channel depth, so that the channel depth before and after each convolution layer is equalized. This allows the identity skip connections to be applied to the model.

### 2.4. Filter Factorization

In [13], simple techniques were used to reduce the parameters in convolution filters. $n \times n$ convolution was decomposed into $n \times 1$ and $1 \times n$ convolution. For example, factorizing $3 \times 3$ convolution filters into $3 \times 1$ and $1 \times 3$ convolution filters saves 33% of parameters [13].

## 3. Proposed Method

In this section we present nonlinear logarithmic filter grouping and residual identity connections in the shallow CNN architecture.

### 3.1. Nonlinear Logarithmic Filter Grouping in Convolution Layers

It is known that human perception is nonlinearly related to external stimuli [16], and is formulated by Weber-Fechner law. The law states that the degree of subjective sensation is proportional to the logarithm of the stimulus intensity, and it can be used to explain auditory and visual systems in humans [24, 25].

We apply this concept of nonlinear perception relation to the aforementioned hierarchical filter groups. In CNNs, convolution filters are supposed to be learned to deliver best responses for input images (or maps). In our method, filters are grouped nonlinearly to assign each filter group with different capacity (e.g., the number of filter or channel size) (refer to Fig. 1). We expect this nonlinear filter grouping could capture the nonlinear nature in the filter distribution (example shown in Fig. 2).

Fig. 2 shows the nonlinear distribution of the spatial frequency of filters in AlexNet. The nonlinear nature may also contain phase, color distributions etc. In this paper, we consider the number of filters along the spatial frequency to show the nonlinear nature of filters. As shown in Fig. 2, CNN filters are not distributed uniformly. Instead, they show a nonlinear distribution. In the first convolution layer of AlexNet, among 96 filters in that convolution layer, the numbers of filters of three filter categories (low, middle and high spatial frequency filter categories) are approximately 53, 28, and 15, respectively.

Typical filter grouping for reducing parameters divides the original full convolution filters into $n$ filter groups with identical channel size [12]. Our nonlinear filter grouping divides the full convolution filters into filter groups with different channel sizes according to nonlinear grouping.

The proposed nonlinear filter grouping uses logarithmic scales with base number 2 to decide the size of each filter group in a convolution layer. A convolution layer has input and output channel depth of $c_{in}$ and $c_{out}$. If the number of filter groups is $n$, then the set of filter shapes of a convolution layer with the logarithmic filter grouping would be

$$\left\{ h \times w \times \frac{c_{in}}{2^i} \times \frac{c_{out}}{2^i} \middle| i = 1, 2, 3, ..., n-2, n-1, n-1 \right\}, \qquad (3)$$

where $h$ and $w$ are height and width size of filters, respectively. When the input and output channel depth are identical, denoted by $c$, all logarithmic filter group sizes in a convolution layer are uniquely defined by $c$ and $n$, for given filter size $h$ and $w$. Channel depth of each filter group would be

$$\left[\frac{c}{2^1}, \frac{c}{2^2}, \frac{c}{2^3}, \dots, \frac{c}{2^{n-2}}, \frac{c}{2^{n-1}}, \frac{c}{2^{n-1}}\right] \quad (4)$$

| Filter grouping scheme | Layer index | Filter group number (*n*) | Filter group size array (*g*) |
|---|---|---|---|
| Logarithmic-4 | Layer 2 | 4 | [64, 32, 16, 16] |
| | Layer 3 | 2 | [128, 128] |
| Logarithmic-8 | Layer 2 | 8 | [64, 32, 16, 8, 4, 2, 1, 1] |
| | Layer 3 | 4 | [128, 64, 32, 32] |
| Logarithmic-16 | Layer 2 | 16 | [32, 16, 16, 8, 8, 8, 8, 8, 4, 4, 4, 4, 4, 2, 1, 1] |
| | Layer 3 | 8 | [128, 64, 32, 16, 8, 4, 2, 2] |

Table 1: Nonlinear logarithmic filter grouping scheme for shallow (3 convolution layers) CNNs for experiment. Note that the logarithmic filter grouping is performed in convolution layer 2 and 3. Layer 2 and layer 3 have channel depths of 128 and 256, respectively.

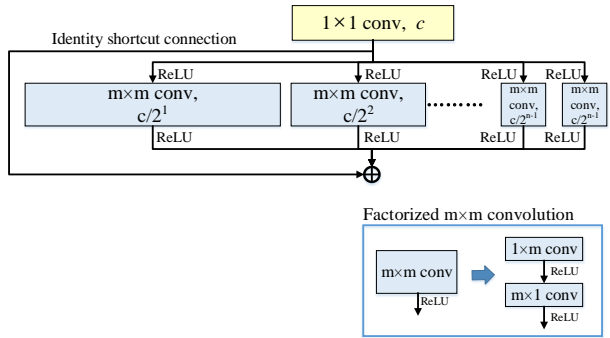

Figure 3: Example of a logarithmic group convolution module with filter group number *n*. *c* indicates the output channel depth. The channel size of each filter group follows the general rule shown in Eq. (4)

We denote Eq. (4) as *g*, which is the filter group size array.

If the value of *n* is too large to divide the filter groups in logarithmic scale (e.g. *n*=16 and *c*=128), we divide the selected filter groups into 2 filter groups with identical size. This process is repeated to create n filter groups. Table 1 shows nonlinear logarithmic filter grouping scheme (with different filter group number *n* and the filter group size array *g*) in shallow CNN networks (3 convolution layers), which are evaluated in the experiments. In Table 1, we show three types of Filter grouping scheme (called Logarithmic-4, Logarithmic-8, and Logarithmic-16) which are deployed to shallow (3 convolution layers) CNNs.

## 3.2 Convolution Module with Logarithmic Filter Grouping

To build efficient shallow CNN, we employ residual identity connection on top of the nonlinear logarithmic filter grouping. When the target feature map and the feature map from identity shortcut have the same dimension, shortcut connection is the identity connection. When the channel depth of both feature maps are different, shortcut connection with 1×1 convolution could be used, but it is proven less effective [2].

We denote 'logarithmic group convolution module' which consists of one 1×1 convolution layer and one *m*×*m* convolution layer to incorporate the residual identity shortcut into the shallow CNN. The 1×1 convolution expands the channel depth of the feature maps before the *m*×*m* convolution to equalize the input and output feature map dimension. This way identity shortcut connection can be applied. In addition, the 1×1 convolution learns a combination of the filter groups of the previous convolution layer [12].

In this paper, without loss of generality, we further reduce filter coefficients by factorization. We factorize the *m*×*m* convolution into 1×*m* and *m*×1 distinct convolutions. Filter grouping is applied to these 1×*m* and *m*×1 convolutions. ReLU activation function is used after each factorized convolution to increase the non-linearity in the network, and all activation functions are kept within the shortcut bypassing the convolution layers. General description of a logarithmic group convolution module is in Fig. 3.

## 4. Experiments

We demonstrate the effectiveness of the proposed logarithmic filter grouping in two different classification tasks. One is facial expression recognition (FER) using Multi-PIE dataset, and the other is object classification using CIFAR-10 dataset.

### 4.1. Evaluation of Proposed Logarithmic Group Convolution Module with Shallow CNN

In order to evaluate the proposed logarithmic filter grouping, we devise compact CNN models which use the proposed logarithmic group convolution module in convolution layers. The shallow network structure we used in experiment was 3 convolution layers-CNN. We applied the filter grouping into 2nd and 3rd convolution layer with the filter grouping scheme seen in Table 1. Fig. 4 is an example of the shallow CNN which employs Logarithmic-8 scheme in Table. 1.

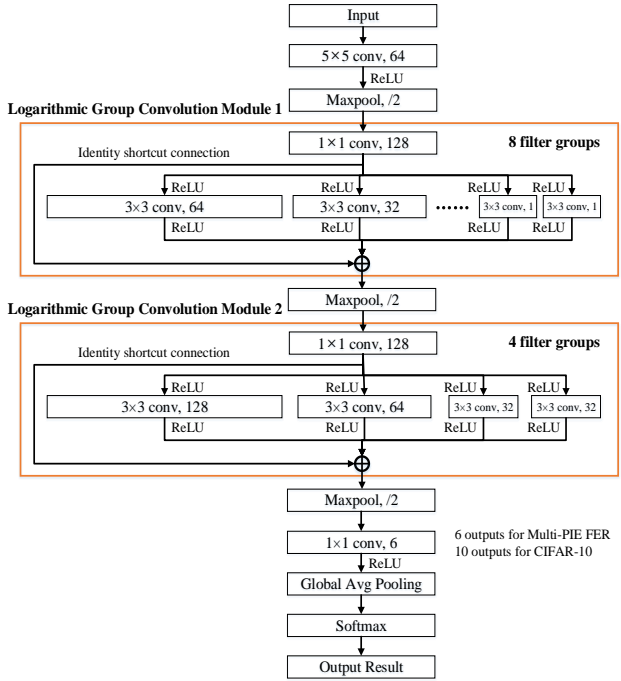

Figure 4: Example of the shallow CNN model used in experiments, which employed the proposed logarithmic group convolution module ($L = 3$, $n = 8$, Logarithmic-8 in Table. 1). Note that factorized convolutions are used for the 3×3 convolutions within the logarithmic group convolution modules.

The CNN is composed of three parts. First part is a convolution layer of filter size 5×5 that expands the channel depth to 64. Second part is convolution modules with filter groups. A 3×3 convolution is used for each module (i.e. 1×3 and 3×1 convolutions), and each module has 128 and 256 output channel depth respectively. Max pooling layers are used to decrease spatial resolution of feature maps. Third part is the classification layer. Instead of a fully connected layer, 1×1 convolution with global average pooling is used as suggested in [26], and a final softmax layer performs classification.

There are three hyperparameters in this network. One is the number of layers $L$. We regard one convolution module as one layer in experiments. Another hyperparameter is the filter group number $n$. Filter group number indicates the degree of grouping in one convolution module. Here, we follow the filter group number decision rule which is shown effective in [12], which is reducing the degree of grouping by half as the depth increases. The other hyperparameter is the filter group size array $g$, which denotes the size of each filter group. This is used to define the nonlinear filter grouping scheme described in Eq. (4).

### 4.1.1. Experiment for FER

**Dataset.** The database we used in the experiments for FER is Multi-PIE dataset, which is originally a face recognition dataset containing more than 750,000 images of 337 subjects in 20 illumination conditions taken from 15 different angles [17]. The dataset contains six facial expressions (neutral, smile, squint, disgust, surprise, scream), and we selected 20,676 images for training and 5,224 images for testing, total 25,900 images. Facial regions were aligned according to the eye centers, cropped and resized to 64 × 64 pixels. For facial alignment, landmark detection method described in [27] was used.

Training images were augmented using rotation ([-7°,-5,-3°,-1°,1°,3°,5°,7°]), translation([-3,3] with 1 pixel step) and scaling ([0.90,1.10] with 0.05 step) The total augmented training images are 640,956 images. 5,224 test images were used for testing the trained network. For data preprocessing, each channel of the cropped input images was normalized to have a zero mean and unit variance.

**Training Details.** In training the shallow network, all networks were trained using Adam optimizer [28]. Parameters for the optimizer were set to default values as suggested in [28] (beta1=0.9, beta2=0.999, epsilon=10$^{-8}$). Learning rate was kept constant to 0.0001 throughout the learning process, and the model was trained for 30 epochs with mini batch size of 128. For loss, standard cross entropy loss was used.

**Evaluation of Multi-PIE FER result.** The results of Multi-PIE FER on the proposed shallow CNN are presented in Table. 2.

The baseline network of the experiment has the same compact structure (i.e., same layer depth, factorized convolution and global average pooling for a compact network) as in Fig. 4, but without filter grouping and residual identity shortcut in convolution modules. In shallow CNN, it is observed that the FER accuracy decreases compared to that of the no filter grouping as the degree of uniform filter grouping increases. This shows that applying filter groups to shallow networks could achieve smaller parameters while degrading the performance. The question is: Can the proposed logarithmic filter grouping reduce the performance degradation?

The results in Table 2 indicate that the networks with the logarithmic filter grouping show better classification accuracy than those with the uniform filter grouping when the filter group numbers are the same. For example, logarithmic-8 achieved about 0.9% higher accuracy compared to uniform-8. This might seem natural as more parameters are used for networks with logarithmic filter groups. However, when comparing logarithmic-8 to uniform-4, we can observe that logarithmic-8 has 53,000 fewer parameters than uniform-4, yet shows modest 0.2% drop in performance. 53,000 parameters take about 10% of the baseline parameter. Regarding uniform-4 has already reduced half of the parameters from the original baseline network, we can still further reduce 10% of the total parameters in the shallow network with a reasonably small loss in performance.

| Filter grouping scheme | Classification accuracy on Multi-PIE FER (%) | Accuracy drop (%) | Total Parameters |
|---|---|---|---|
| Uniform-4 w/o shortcut | 86.54 | 0.48 | 268,480 |
| Uniform-8 w/o shortcut | 85.18 | 1.84 | 157,888 |
| Uniform-16 w/o shortcut | 84.67 | 2.35 | 102,592 |
| Uniform-4 | 86.81 | 0.21 | 268,480 |
| Uniform-8 | 85.70 | 1.32 | 157,888 |
| Uniform-16 | 85.13 | 1.89 | 102,592 |
| Logarithmic-4 | **86.98** | **0.04** | **277,696** |
| Logarithmic-8 | 86.59 | 0.43 | 215,236 |
| Logarithmic-16 | 86.20 | 0.82 | 190,036 |
| No filter grouping (baseline) | 87.02 | - | 543,616 |

Table 2: Total parameter size of the compact CNNs and associated classification accuracy (%) on Multi-PIE FER dataset. Uniform-*n* uses uniform filter grouping with filter group number *n* in the convolution modules and Logarithmic-*n* uses nonlinear logarithmic group convolution modules with the grouping scheme defined in Table 1. Uniform-*n* w/o shortcut has the same network structure as Uniform-*n*, but without residual identity shortcut. Among filter grouping schemes, the best performance is indicated in **bold**, and the best parameter efficiency is in blue. Note that logarithmic-8 shows small 0.2% accuracy drop compared to uniform-4, while having 53,000 (20%) less parameters than uniform-4. Also, logarithmic-4 presents similar performance to the baseline while having 50% less parameters compared to the baseline.

Logarithmic-4 also shows improved accuracy and it presents similar performance compared to the baseline while having 50% less parameters. Logarithmic-4 and 16 both showed improved performance, but considering the number of parameters the accuracy increase is not as large as logarithmic-8. Logarithmic-4 uses the logarithmic filter grouping only in layer 2 according to the grouping scheme we defined, and this might not be enough to reflect the nonlinearity in all filters in the network. Adding to this, it can be interpreted that the filter nonlinearity of the shallow network trained with Multi-PIE FER dataset is best represented by the filter grouping scheme of logarithmic-8.

The residual identity shortcut is also shown to be effective in the shallow CNN. For all networks with uniform filter grouping, the accuracy increased when convolution module with identity shortcut was used.

### 4.1.2. Experiment for Object Classification

**Dataset.** CIFAR-10 dataset [18] contains color images of different objects, with 32×32 pixels each. There are 10 classes, and the training and test sets consist of 50,000 and 10,000 images. Each class has 6,000 images. We followed the standard data augmentation scheme used in [7, 20, 26, 29, 30] for training: images are padded by 4 pixels on each side, and a random 32×32 crop is sampled from the padded image or its horizontal flip. For testing, we used the 10,000 test images without alterations.

**Training Details.** Adam optimizer was used with the same parameters as in FER experiment with different learning rate. Learning rate was kept constant to 0.001 until 100 epochs, and halved at 100 epochs, dropped down to 0.0001 at 140 epochs, and halved at 160 epochs and kept constant up to 180 epochs. Mini batch size of 128 was used and standard cross entropy loss was also used.

**Evaluation of CIFAR-10 result:** The results of CIFAR-10 object classification experiment displayed similar trend shown in FER experiment. The results are presented in Table. 3.

Similar to the Multi-PIE FER result, the classification accuracy of CIFAR-10 drops as the degree of uniform filter grouping increases. Also, as shown before, the residual identity shortcut redeems the decreased accuracy due to filter grouping. The increase in accuracy is about 0.3% for all uniform-4, 8 and 16 networks which is similar amount compared to the previous experiment.

Overall, networks with the logarithmic filter grouping outperform networks with the uniform filter grouping, and the general improvement is larger than that of Multi-PIE FER. One noticeable observation is that the accuracy of logarithmic-8 is even better than uniform-4, while having 53,000 fewer parameters. Also, logarithmic-8 has slightly better performance compared to logarithmic-4. This result supports the idea that whilst having less parameter than logarithmic-4, logarithmic-8 better represents the nonlinear nature in the network trained with CIFAR-10.

| Filter grouping scheme | Classification accuracy on CIFAR-10 (%) | Accuracy drop (%) | Total Parameters |
|---|---|---|---|
| Uniform-4 w/o shortcut | 85.27 | 1.79 | 269,504 |
| Uniform-8 w/o shortcut | 84.24 | 2.82 | 158,912 |
| Uniform-16 w/o shortcut | 83.19 | 3.87 | 103,616 |
| Uniform-4 | 85.53 | 1.53 | 269,504 |
| Uniform-8 | 84.54 | 2.52 | 158,912 |
| Uniform-16 | 83.97 | 3.09 | 103,616 |
| Logarithmic-4 | 85.62 | 1.44 | 278,720 |
| Logarithmic-8 | **85.79** | **1.27** | **216,260** |
| Logarithmic-16 | 85.09 | 1.97 | 191,060 |
| No filter grouping (baseline) | 87.06 | - | 544,640 |

Table 3: Total parameter size of the compact CNNs and associated classification accuracy (%) on CIFAR-10 dataset. Each network has the same structure as described in Table. 2. Number of parameters of each network is increased by 1024 due to the additional 4 classes in CIFAR-10 compared to Multi-PIE FER. Among filter grouping schemes, the best performance is indicated in **bold**, and the best parameter efficiency is in blue. Note that logarithmic-8 shows better performance than uniform-4, while having 20% less parameters than uniform-4.

## 5. Conclusion

We proposed a new filter grouping method which adapts the nonlinear logarithmic filter grouping. The logarithmic filter grouping divides the convolution layer filters in logarithmic sizes, and this grouping scheme reflects the nonlinear nature in filter distribution. To apply the proposed method to shallow CNN structure, we devised a shallow network with logarithmic group convolution modules. This module allows us to use both logarithmic filter grouping and residual identity shortcut in the shallow CNN.

To validate the effectiveness of our method in shallow networks, the suggested shallow CNN with three different logarithmic filter grouping schemes were tested with Multi-PIE FER and CIFAR-10 object classification. The results showed that all networks with the logarithmic filter grouping schemes outperformed the same networks with uniform filter grouping in both experiments. From the parameter point of view, the logarithmic filter grouping could further reduce the number of parameters while maintaining or enhancing the performance compared to the uniform filter grouping. The residual identity shortcut is also shown effective in the shallow CNN, presenting slight increase in performance compared to networks with no identity shortcuts.

The proposed logarithmic filter grouping and shallow CNN can help reducing network sizes for mobile applications with constrained conditions. Further work on deciding different nonlinear filter grouping schemes may help increasing the efficiency of shallow CNNs even more. As a future work, we are going to apply the proposed logarithmic filter grouping to deep networks to show its usefulness in parameter reduction for deep networks.